\documentclass{article}

\usepackage{times}
\usepackage{graphicx}
\usepackage{subfigure} 
\usepackage{natbib}
\usepackage{algorithm}
\usepackage{algorithmic}
\usepackage{hyperref}
\usepackage{amsmath}
\usepackage{amssymb}
\usepackage{nicefrac}
\usepackage{appendix}
\usepackage{tikz}

\usepackage[accepted]{icml2016}

\newcommand\floor[1]{\lfloor#1\rfloor}

\newcommand\W[1]{W_\text{#1}}
\newcommand\B[1]{B_\text{#1}}
\icmltitlerunning{Deep Voice: Real-time Neural TTS}

\begin{document} 

\renewcommand*{\thefootnote}{\fnsymbol{footnote}}
\twocolumn[
\icmltitle{Deep Voice: Real-time Neural Text-to-Speech}

\icmlauthor{Sercan \"{O}.  Ar{\i}k\textsuperscript{\textdagger}}{sercanarik@baidu.com}
\icmlauthor{Mike Chrzanowski\textsuperscript{\textdagger}}{mikechrzanowski@baidu.com}
\icmlauthor{Adam Coates\textsuperscript{\textdagger}}{adamcoates@baidu.com}
\icmlauthor{Gregory Diamos\textsuperscript{\textdagger}}{gregdiamos@baidu.com}
\icmlauthor{Andrew Gibiansky\textsuperscript{\textdagger}}{gibianskyandrew@baidu.com}
\icmlauthor{Yongguo Kang\textsuperscript{\textdagger}}{kangyongguo@baidu.com}   
\icmlauthor{Xian Li\textsuperscript{\textdagger}}{lixian05@baidu.com}           
\icmlauthor{John Miller\textsuperscript{\textdagger}}{millerjohn@baidu.com}
\icmlauthor{Andrew Ng\textsuperscript{\textdagger}}{andrewng@baidu.com}
\icmlauthor{Jonathan Raiman\textsuperscript{\textdagger}}{jonathanraiman@baidu.com}
\icmlauthor{Shubho Sengupta\textsuperscript{\textdagger}}{ssengupta@baidu.com}
\icmlauthor{Mohammad Shoeybi\textsuperscript{\textdagger}}{mohammad@baidu.com}
\icmladdress{Baidu Silicon Valley Artificial Intelligence Lab, 1195 Bordeaux Dr. Sunnyvale, CA 94089}

\icmlkeywords{speech generation, deep voice, deep learning, text-to-speech, tts, deep neural networks, machine learning, convolutional neural networks, recurrent neural networks}

\vskip 0.2in
]
\footnotetext[2]{Authors are listed alphabetically by last name.}
\renewcommand*{\thefootnote}{\arabic{footnote}}
\newcommand\forarxiv{yes}

\ifdefined\forarxiv
\newcommand\aftersectionvfill{\vspace{0in}}
\else
\renewcommand{\baselinestretch}{0.95}
\newcommand\aftersectionvfill{\vspace{-0.13in}}
\setlength{\parskip}{4pt}
\fi

\begin{abstract} 
   We present Deep Voice, a production-quality text-to-speech system constructed entirely from deep neural networks. Deep Voice lays the groundwork for truly end-to-end neural speech synthesis. The system comprises five major building blocks: a segmentation model for locating phoneme boundaries, a grapheme-to-phoneme conversion model, a phoneme duration prediction model, a fundamental frequency prediction model, and an audio synthesis model. For the segmentation model, we propose a novel way of performing phoneme boundary detection with deep neural networks using connectionist temporal classification (CTC) loss. For the audio synthesis model, we implement a variant of WaveNet that requires fewer parameters and trains faster than the original. By using a neural network for each component, our system is simpler and more flexible than traditional text-to-speech systems, where each component requires laborious feature engineering and extensive domain expertise. Finally, we show that inference with our system can be performed faster than real time and describe optimized WaveNet inference kernels on both CPU and GPU that achieve up to 400x speedups over existing implementations.
\end{abstract}

\aftersectionvfill\section{Introduction}
\label{introduction}

Synthesizing artificial human speech from text, commonly known as text-to-speech (TTS), is an essential component in many applications such as speech-enabled devices, navigation systems, and accessibility for the visually-impaired. Fundamentally, it allows human-technology interaction without requiring visual interfaces. Modern TTS systems are based on complex, multi-stage processing pipelines, each of which may rely on hand-engineered features and heuristics. Due to this complexity, developing new TTS systems can be very labor intensive and difficult.

Deep Voice is inspired by traditional text-to-speech pipelines and adopts the same structure, while replacing all components with neural networks and using simpler features: first we convert text to phoneme and then use an audio synthesis model to convert linguistic features into speech \cite{taylor2009tts_book}. Unlike prior work (which uses hand-engineered features such as spectral envelope, spectral parameters, aperiodic parameters, etc.), our only features are phonemes with stress annotations, phoneme durations, and fundamental frequency (F0). This choice of features makes our system more readily applicable to new datasets, voices, and domains without any manual data annotation or additional feature engineering. We demonstrate this claim by retraining our entire pipeline without any hyperparameter changes on an entirely new dataset that contains solely audio and unaligned textual transcriptions and generating relatively high quality speech. In a conventional TTS system this adaptation requires days to weeks of tuning, whereas Deep Voice allows you to do it in only a few hours of manual effort and the time it takes models to train.

Real-time inference is a requirement for a production-quality TTS system; without it, the system is unusable for most applications of TTS. Prior work has demonstrated that a WaveNet \cite{van2016wavenet} can generate close to human-level speech. However, WaveNet inference poses a daunting computational problem due to the high-frequency, autoregressive nature of the model, and it has been hitherto unknown whether such models can be used in a production system. We answer this question in the affirmative and demonstrate efficient, faster-than-real-time WaveNet inference kernels that produce high-quality 16 kHz audio and realize a 400X speedup over previous WaveNet inference implementations \cite{paine2016fast}.

\aftersectionvfill\section{Related Work}
\label{related_work}

Previous work uses neural networks as substitutes for several TTS system components, including grapheme-to-phoneme conversion models \cite{rao2015grapheme, yao2015sequence}, phoneme duration prediction models \cite{zen2015unidirectional}, fundamental frequency prediction models \cite{pascual1000multi, ronanki2016template}, and audio synthesis models \cite{van2016wavenet, mehri2016samplernn}. Unlike Deep Voice, however, none of these systems solve the entire problem of TTS and many of them use specialized hand-engineered features developed specifically for their domain.

Most recently, there has been a lot of work in parametric audio synthesis, notably WaveNet, SampleRNN, and Char2Wav \cite{van2016wavenet, mehri2016samplernn, char2wav}. While WaveNet can be used for both conditional and unconditional audio generation, SampleRNN is only used for unconditional audio generation. Char2Wav extends SampleRNN with an attention-based phoneme duration model and the equivalent of an F0 prediction model, effectively providing local conditioning information to a SampleRNN-based vocoder.

Deep Voice differs from these systems in several key aspects that notably increase the scope of the problem. First, Deep Voice is completely standalone; training a new Deep Voice system does not require a pre-existing TTS system, and can be done from scratch using a dataset of short audio clips and corresponding textual transcripts. In contrast, reproducing either of the aforementioned systems requires access and understanding of a pre-existing TTS system, because they use features from another TTS system either at training or inference time.

Second, Deep Voice minimizes the use of hand-engineered features; it uses one-hot encoded characters for grapheme to phoneme conversion, one-hot encoded phonemes and stresses, phoneme durations in milliseconds, and normalized log fundamental frequency that can be computed from waveforms using any F0 estimation algorithm. All of these can easily be obtained from audio and transcripts with minimal effort. In contrast, prior works use a much more complex feature representation, that effectively makes reproducing the system impossible without a pre-existing TTS system. WaveNet uses several features from a TTS system \cite{zen2013nnsynthesis}, that include values such as the number of syllables in a word, position of syllables in the phrase, position of the current frame in the phoneme, and dynamic features of the speech spectrum like spectral and excitation parameters, as well as their time derivatives. Char2Wav relies on vocoder features from the WORLD TTS system \cite{morise2016world} for pre-training their alignment module which include F0, spectral envelope, and aperiodic parameters.

Finally, we focus on creating a production-ready system, which \emph{requires} that our models run in real-time for inference. Deep Voice can synthesize audio in fractions of a second, and offers a tunable trade-off between synthesis speed and audio quality. In contrast, previous results with WaveNet require several minutes of runtime to synthesize one second of audio. We are unaware of similar benchmarks for SampleRNN, but the 3-tier architecture as described in the original publication requires approximately 4-5X as much compute during inference as our largest WaveNet models, so running the model in real-time may prove challenging.

\aftersectionvfill\section{TTS System Components}
\label{models}

\begin{figure*}[!ht]
    \centering
    \includegraphics[width=0.8\textwidth]{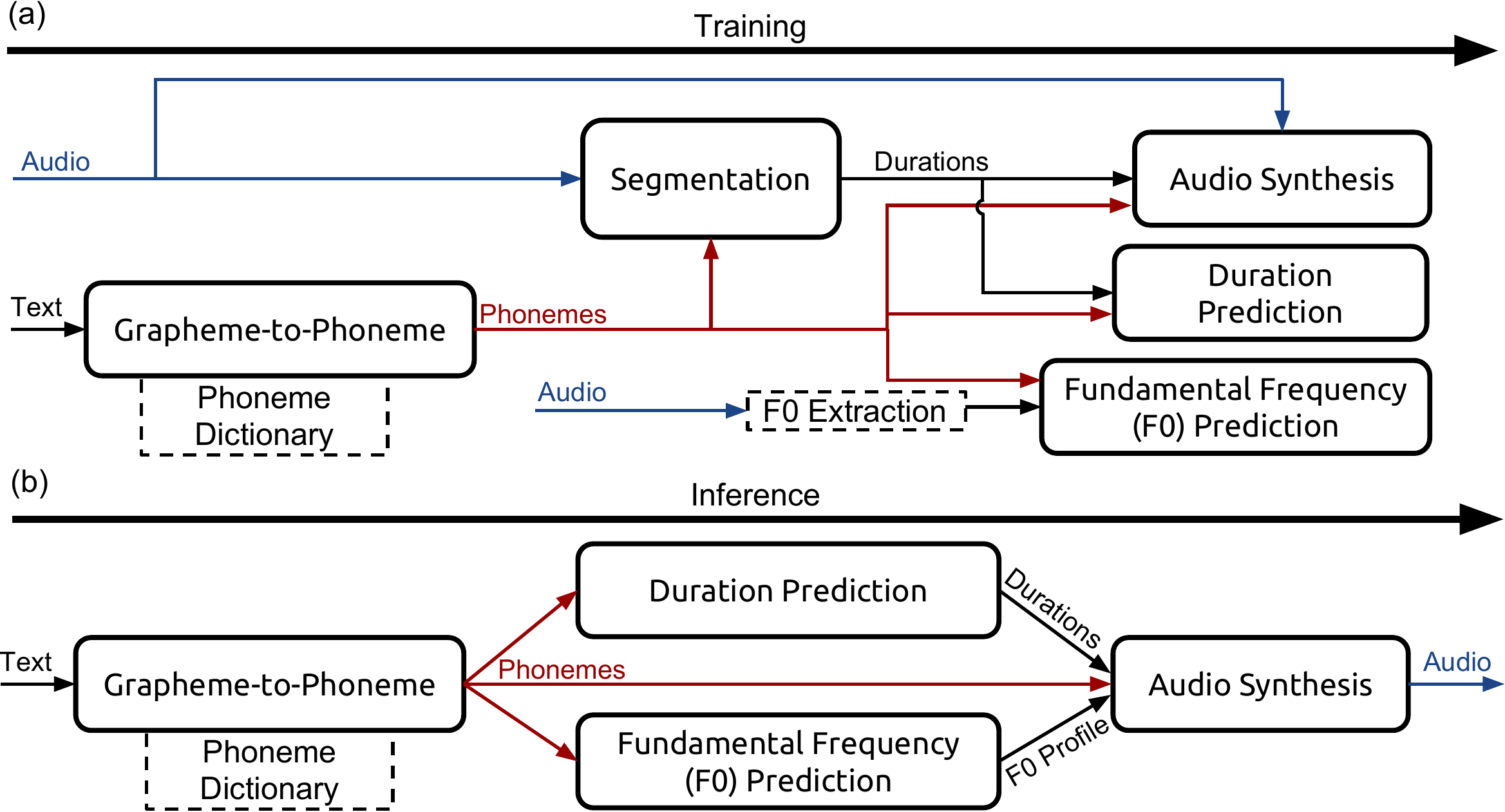}
    \caption{System diagram depicting (a) training procedure and (b) inference procedure, with inputs on the left and outputs on the right. In our system, the duration prediction model and the F0 prediction model are performed by a single neural network trained with a joint loss. The grapheme-to-phoneme model is used as a fallback for words that are not present in a phoneme dictionary, such as CMUDict. Dotted lines denote non-learned components.}
    \label{fig:system-diagram}
\end{figure*}

As shown in Fig.~\ref{fig:system-diagram}, the TTS system consists of five major building blocks:

\begin{itemize}
    \itemsep0em 
    \item The \textbf{grapheme-to-phoneme model} converts from written text (English characters) to
        phonemes (encoded using a phonemic alphabet such as ARPABET).
    \item The \textbf{segmentation model} locates phoneme boundaries in the voice dataset. Given an
        audio file and a phoneme-by-phoneme transcription of the audio, the segmentation model
        identifies where in the audio each phoneme begins and ends.
    \item The \textbf{phoneme duration model} predicts the temporal duration of every phoneme in a phoneme sequence (an utterance).
    \item The \textbf{fundamental frequency model} predicts whether a phoneme is voiced. If it is, the model predicts the fundamental frequency (F0) throughout the phoneme's duration.
    \item The \textbf{audio synthesis model} combines the outputs of the grapheme-to-phoneme, phoneme duration, and fundamental frequency prediction models and synthesizes audio at a high sampling rate, corresponding to the desired text.
\end{itemize}

During inference, text is fed through the grapheme-to-phoneme model or a phoneme dictionary to generate phonemes. Next, the phonemes are provided as inputs to the phoneme duration model and F0 prediction model to assign durations to each phoneme and generate an F0 contour. Finally, the phonemes, phoneme durations, and F0 are used as local conditioning input features to the audio synthesis model, which generates the final utterance.

Unlike the other models, the segmentation model is not used during inference. Instead, it is used to annotate the training voice data with phoneme boundaries. The phoneme boundaries imply durations, which can be used to train the phoneme duration model. The audio, annotated with phonemes and phoneme durations as well as fundamental frequency, is used to train the audio synthesis model.

In the following sections, we describe all the building blocks in detail.

\subsection{Grapheme-to-Phoneme Model}

Our grapheme-to-phoneme model is based on the encoder-decoder architecture developed by \cite{yao2015sequence}. However, we use a multi-layer bidirectional encoder with a gated recurrent unit (GRU) nonlinearity and an equally deep unidirectional GRU decoder \cite{chung2014empirical}. The initial state of every decoder layer is initialized to the final hidden state of the corresponding encoder forward layer. The architecture is trained with teacher forcing and decoding is performed using beam search. We use 3 bidirectional layers with 1024 units each in the encoder and 3 unidirectional layers of the same size in the decoder and a beam search with a width of 5 candidates. During training, we use dropout with probability 0.95 after each recurrent layer.

For training, we use the Adam optimization algorithm with $\beta_1=0.9, \beta_2=0.999, \varepsilon=10^{-8}$, a batch size of 64, a learning rate of $10^{-3}$, and an annealing rate of 0.85 applied every 1000 iterations \cite{ADAM}. 

\subsection{Segmentation Model}

Our segmentation model is trained to output the alignment between a given utterance and a sequence of target phonemes. This task is similar to the problem of aligning speech to written output in speech recognition. In that domain, the connectionist temporal classification (CTC) loss function has been shown to focus on character alignments to learn a mapping between sound and text \cite{graves2006ctc}. We adapt the convolutional recurrent neural network architecture from a state-of-the-art speech recognition system \cite{amodei2015deep} for phoneme boundary detection.

A network trained with CTC to generate sequences of phonemes will produce brief peaks for every output phoneme. Although this is sufficient to roughly align the phonemes to the audio, it is insufficient to detect precise phoneme boundaries. To overcome this, we train to predict sequences of phoneme \emph{pairs} rather than single phonemes. The network will then tend to output phoneme pairs at timesteps close to the boundary between two phonemes in a pair. 

To illustrate our label encoding, consider the string ``Hello!''. To convert this to a sequence of phoneme pair labels, convert the utterance to phonemes (using a pronunciation dictionary such as CMUDict or a grapheme-to-phoneme model) and pad the phoneme sequence on either end with the silence phoneme to  get ``sil HH EH L OW sil''. Finally, construct consecutive phoneme pairs and get ``(sil, HH), (HH, EH), (EH, L), (L, OW), (OW, sil)''.

Input audio is featurized by computing 20 Mel-frequency cepstral coefficients (MFCCs) with a ten millisecond stride. On top of the input layer, there are two convolution layers (2D convolutions in time and frequency), three bidirectional recurrent GRU layers, and finally a softmax output layer. The convolution layers use kernels with unit stride, height nine (in frequency bins), and width five (in time) and the recurrent layers use 512 GRU cells (for each direction). Dropout with a probability of 0.95 is applied after the last convolution and recurrent layers. To compute the phoneme-pair error rate (PPER), we decode using beam search. To decode phoneme boundaries, we perform a beam search with width 50 with the constraint that neighboring phoneme pairs overlap by at least one phoneme and keep track of the positions in the utterance of each phoneme pair.

For training, we use the Adam optimization algorithm with $\beta_1=0.9, \beta_2=0.999, \varepsilon=10^{-8}$, a batch size of 128, a learning rate of $10^{-4}$, and an annealing rate of 0.95 applied every 500 iterations \cite{ADAM}. 

\subsection{Phoneme Duration and Fundamental Frequency Model}

We use a single architecture to jointly predict phoneme duration and time-dependent fundamental frequency. The input to the model is a sequence of phonemes with stresses, with each phoneme and stress being encoded as a one-hot vector. The architecture comprises two fully connected layers with 256 units each followed by two unidirectional recurrent layers with 128 GRU cells each and finally a fully-connected output layer. Dropout with a probability of 0.8 is applied after the initial fully-connected layers and the last recurrent layer.

The final layer produces three estimations for every input phoneme: the phoneme duration, the probability that the phoneme is voiced (i.e. has a fundamental frequency), and 20 time-dependent F0 values, which are sampled uniformly over the predicted duration.

The model is optimized by minimizing a joint loss that combines phoneme duration error, fundamental frequency error, the negative log likelihood of the probability that the phoneme is voiced, and a penalty term proportional to the absolute change of F0 with respect to time to impose smoothness. The specific functional form of the loss function is described in Appendix~\ref{app:phonememodel}.\ifdefined\forarxiv\else\footnote{Appendices and audio samples are available \ifdefined\isaccepted
    on Arxiv.
\else
    as supplementary material.
\fi
}\fi

For training, we use the Adam optimization algorithm with $\beta_1=0.9, \beta_2=0.999, \varepsilon=10^{-8}$, a batch size of 128, a learning rate of $3\times10^{-4}$, and an annealing rate of 0.9886 applied every 400 iterations \cite{ADAM}.

\subsection{Audio Synthesis Model}

Our audio synthesis model is a variant of WaveNet. WaveNet consists of a conditioning network, which upsamples linguistic features to the desired frequency, and an autoregressive network, which generates a probability distribution $\mathbb P(y)$ over discretized audio samples $y \in \{0, 1, \ldots, 255\}$. We vary the number of layers $\ell$, the number of residual channels $r$ (dimension of the hidden state of every layer), and the number of skip channels $s$ (the dimension to which layer outputs are projected prior to the output layer).

WaveNet consists of an upsampling and conditioning network, followed by $\ell$ $2\times1$ convolution layers with $r$ residual output channels and gated $\mathrm{tanh}$ nonlinearities. We break the convolution into two matrix multiplies per timestep with $\W{prev}$ and $\W{cur}$. These layers are connected with residual connections. The hidden state of every layer is concatenated to an $\ell r$ vector and projected to $s$ skip channels with $\W{skip}$, followed by two layers of $1\times1$ convolutions (with weights $\W{relu}$ and $\W{out}$) with $\mathrm{relu}$ nonlinearities.

WaveNet uses transposed convolutions for upsampling and conditioning. We find that our models perform better, train faster, and require fewer parameters if we instead first encode the inputs with a stack of bidirectional quasi-RNN (QRNN) layers \cite{bradbury2016quasi} and then perform upsampling by repetition to the desired frequency.

Our highest-quality final model uses $\ell=40$ layers, $r=64$ residual channels, and $s=256$ skip channels. For training, we use the Adam optimization algorithm with $\beta_1=0.9, \beta_2=0.999, \varepsilon=10^{-8}$, a batch size of 8, a learning rate of $10^{-3}$, and an annealing rate of 0.9886 applied every 1,000 iterations \cite{ADAM}.

Please refer to Appendix~\ref{app:wavenet-architecture} for full details of our WaveNet architecture and the QRNN layers we use.

\section{Results}
\label{results}

We train our models on an internal English speech database containing approximately 20 hours of speech data segmented into 13,079 utterances. In addition, we present audio synthesis results for our models trained on a subset of the Blizzard 2013 data \cite{prahallad2013blizzard}. Both datasets are spoken by a professional female speaker.

All of our models are implemented using the TensorFlow framework \cite{tensorflow2015-whitepaper}.

\subsection{Segmentation Results}

We train on 8 TitanX Maxwell GPUs, splitting each batch equally among the GPUs and using a ring all-reduce to average gradients computed on different GPUs, with each iteration taking approximately 1300 milliseconds. After approximately 14,000 iterations, the model converges to a phoneme pair error rate of 7\%. We also find that phoneme boundaries do not have to be precise, and randomly shifting phoneme boundaries by 10-30 milliseconds makes no difference in the audio quality, and so suspect that audio quality is insensitive to the phoneme pair error rate past a certain point.

\subsection{Grapheme-to-Phoneme Results}

We train a grapheme-to-phoneme model on data obtained from CMUDict \cite{weir2008cmudict}. We strip out all words that do not start with a letter, contain numbers, or have multiple pronunciations, which leaves 124,978 out of the original 133,854 grapheme-phoneme sequence pairs.

We train on a single TitanX Maxwell GPU with each iteration taking approximately 150 milliseconds. After approximately 20,000 iterations, the model converges to a phoneme error rate of 5.8\% and a word error rate of 28.7\%, which are on par with previous reported results \cite{yao2015sequence}. Unlike prior work, we do not use a language model during decoding and do not include words with multiple pronunciations in our data set.

\subsection{Phoneme Duration and Fundamental Frequency Results}

We train on a single TitanX Maxwell GPU with each iteration taking approximately 120 milliseconds. After approximately 20,000 iterations, the model converges to a mean absolute error of 38 milliseconds (for phoneme duration) and 29.4 Hz (for fundamental frequency).

\subsection{Audio Synthesis Results}
\label{audio_syn_results}

We divide the utterances in our audio dataset into one second chunks with a quarter second of context for each chunk, padding each utterance with a quarter second of silence at the beginning. We filter out chunks that are predominantly silence and end up with 74,348 total chunks.

We trained models with varying depth, including 10, 20, 30, and 40 layers in the residual layer stack. We find that models below 20 layers result in poor quality audio. The 20, 30, and 40 layer models all produce high quality recognizable speech, but the 40 layer models have less noise than the 20 layer models, which can be detected with high-quality over-ear headphones.

Previous work has emphasized the importance of receptive field size in determining model quality. Indeed, the 20 layer models have half the receptive field as the 40 layer models. However, when run at 48 kHz, models with 40 layers have only 83 milliseconds of receptive field, but still generate high quality audio. This suggests the receptive field of the 20 layer models is sufficient, and we conjecture the difference in audio quality is due to some other factor than receptive field size.

We train on 8 TitanX Maxwell GPUs with one chunk per GPU, using a ring allreduce to average gradients computed on different GPUs. Each iteration takes approximately 450 milliseconds. Our model converges after approximately 300,000 iterations. We find that a single 1.25s chunk is sufficient to saturate the compute on the GPU and that batching does not increase training efficiency.

As is common with high-dimensional generative models \cite{theis2015note}, model loss is somewhat uncorrelated with perceptual quality of individual samples. While models with unusually high loss sound distinctly noisy, models that optimize below a certain threshold do not have a loss indicative of their quality. In addition, changes in model architecture (such as depth and output frequency) can have a significant impact on model loss while having a small effect on audio quality.

To estimate perceptual quality of the individual stages of our TTS pipeline, we crowdsourced mean opinion score (MOS) ratings (ratings between one and five, higher values being better) from Mechanical Turk using the CrowdMOS toolkit and methodology \cite{ribeiro2011crowdmos}. In order to separate the effect of the audio preprocessing, the WaveNet model quality, and the phoneme duration and fundamental frequency model quality, we present MOS scores for a variety of utterance types, including synthesis results where the WaveNet inputs (duration and F0) are extracted from ground truth audio rather than synthesized by other models. The results are presented in Table~\ref{tab:mos}. We purposefully include ground truth samples in every batch of samples that raters evaluate to highlight the delta from human speech and allow raters to distinguish finer grained differences between models; the downside of this approach is that the resulting MOS scores will be significantly lower than if raters are presented \emph{only} with synthesized audio samples.

First of all, we find a significant drop in MOS when simply downsampling the audio stream from 48 kHz to 16 kHz, especially in combination with $\mu$-law companding and quantization, likely because a 48 kHz sample is presented to the raters as a baseline for a 5 score, and a low quality noisy synthesis result is presented as a 1. When used with ground truth durations and F0, our models score highly, with the 95\% confidence intervals of our models intersecting those of the ground truth samples. However, using synthesized frequency reduces the MOS, and further including synthesized durations reduces it significantly. We conclude that the main barrier to progress towards natural TTS lies with duration and fundamental frequency prediction, and our systems have not meaningfully progressed past the state of the art in that regard. Finally, our best models run slightly slower than real-time (see Table~\ref{tab:inference-speed}), so we demonstrate that synthesis quality can be traded for inference speed by adjusting model size by obtaining scores for models that run 1X and 2X faster than real-time.

We also tested WaveNet models trained on the full set of features from the original WaveNet publication, but found no perceptual difference between those models and models trained on our reduced feature set.

\begin{table*}[ht]
    \centering
    \begin{tabular}{|l|c|l|l|}
        \hline
\textbf{Type}                             & \textbf{Model Size}   & \textsc{MOS}$\pm$CI  \\ \hline
Ground Truth (48 kHz)                     & None                  & $4.75\pm0.12$\ \\ \hline
Ground Truth                              & None                  & $4.45\pm0.16$ \\ \hline
Ground Truth (companded and expanded)     & None                  & $4.34\pm0.18$ \\ \hline
Synthesized                               & $\ell=40,r=64,s=256$  & $3.94\pm0.26$ \\ \hline
Synthesized (48 kHz)                      & $\ell=40,r=64,s=256$  & $3.84\pm0.24$ \\ \hline
Synthesized (Synthesized F0)              & $\ell=40,r=64,s=256$  & $2.76\pm0.31$ \\ \hline
Synthesized (Synthesized Duration and F0) & $\ell=40,r=64,s=256$  & $2.00\pm0.23$ \\ \hline
Synthesized (2X real-time inference)       & $\ell=20,r=32,s=128$  & $2.74\pm0.32$ \\ \hline
Synthesized (1X real-time inference)       & $\ell=20,r=64,s=128$  & $3.35\pm0.31$ \\ \hline
    \end{tabular}
    \caption{Mean Opinion Scores (MOS) and 95\% confidence intervals (CIs) for utterances. This MOS score is a relative MOS score obtained by showing raters the same utterance across all the model types (which encourages comparative rating and allows the raters to distinguish finer grained differences). Every batch of samples also includes the ground truth 48 kHz recording, which makes all our ratings comparative to natural human voices. 474 ratings were collected for every sample. Unless otherwise mentioned, models used phoneme durations and F0 extracted from the ground truth, rather than synthesized by the duration prediction and frequency prediction models, as well as a 16384 Hz audio sampling rate.}
    \label{tab:mos}
\end{table*}

\begin{table*}[ht]
    \centering
    \begin{tabular}{|c|c|c|c|c|}
        \hline
\textbf{Model}         & \textbf{Platform} & \textbf{Data Type} & \textbf{Number of Threads} & \textbf{Speed-up Over Real-time} \\ \hline
$\ell=20, r=32, s=128$ & \textsc{CPU}      & \texttt{float32}   & 6                 & \textbf{2.7}  \\ \hline
$\ell=20, r=32, s=128$ & \textsc{CPU}      & \texttt{float32}   & 2                 & \textbf{2.05} \\ \hline
$\ell=20, r=64, s=128$ & \textsc{CPU}      & \texttt{int16}     & 2                 & \textbf{1.2}  \\ \hline
$\ell=20, r=64, s=128$ & \textsc{CPU}      & \texttt{float32}   & 6                 & \textbf{1.11} \\ \hline
$\ell=20, r=64, s=128$ & \textsc{CPU}      & \texttt{float32}   & 2                 & 0.79          \\ \hline
$\ell=40, r=64, s=256$ & \textsc{CPU}      & \texttt{int16}     & 2                 & 0.67          \\ \hline
$\ell=40, r=64, s=256$ & \textsc{CPU}      & \texttt{float32}   & 6                 & 0.61          \\ \hline
$\ell=40, r=64, s=256$ & \textsc{CPU}      & \texttt{float32}   & 2                 & 0.35          \\ \hline\hline
$\ell=20, r=32, s=128$ & \textsc{GPU}      & \texttt{float32}   & N/A               & 0.39          \\ \hline
$\ell=20, r=64, s=128$ & \textsc{GPU}      & \texttt{float32}   & N/A               & 0.29          \\ \hline
$\ell=40, r=32, s=128$ & \textsc{GPU}      & \texttt{float32}   & N/A               & 0.23          \\ \hline
$\ell=40, r=64, s=128$ & \textsc{GPU}      & \texttt{float32}   & N/A               & 0.17          \\ \hline
    \end{tabular}
    \caption{CPU and GPU inference kernel benchmarks for different models in float32 and int16. At least one main and one auxiliary thread were used for all CPU kernels. These kernels operate on a single utterance with no batching. CPU results are from a Intel Xeon E5-2660 v3 Haswell processor clocked at 2.6 GHz and GPU results are from a GeForce GTX Titan X Maxwell GPU.}
    \label{tab:inference-speed}
\end{table*}

\subsection{Blizzard Results}
To demonstrate the flexibility of our system, we retrained all of our models with identical hyperparameters on the Blizzard 2013 dataset \cite{prahallad2013blizzard}. For our experiments, we used a 20.5 hour subset of the dataset segmented into 9,741 utterances. We evaluated the model using the procedure described in Section \ref{audio_syn_results}, which encourages raters to compare synthesized audio directly with the ground truth. On the held out set, 16 kHz companded and expanded audio receives a MOS score of $4.65\pm0.13$, while our synthesized audio received a MOS score of $2.67\pm0.37$.

\aftersectionvfill\section{Optimizing Inference}
\label{inference}

Although WaveNet has shown promise in generating high-quality synthesized speech, initial experiments reported generation times of many minutes or hours for short utterances. WaveNet inference poses an incredibly challenging computational problem due to the high-frequency, autoregressive nature of the model, which requires orders of magnitude more timesteps than traditional recurrent neural networks. When generating audio, a single sample must be generated in approximately 60 $\mu$s (for 16 kHz audio) or 20 $\mu$s (for 48 kHz audio). For our 40 layer models, this means that a single layer (consisting of several matrix multiplies and nonlinearities) must complete in approximately 1.5 $\mu$s. For comparison, accessing a value that resides in main memory on a CPU can take 0.1 $\mu$s. In order to perform inference at real-time, we must take great care to never recompute any results, store the entire model in the processor cache (as opposed to main memory), and optimally utilize the available computational units. These same techniques could be used to accelerate image synthesis with PixelCNN \cite{oord2016pixel} to fractions of a second per image.

Synthesizing one second of audio with our 40 layer WaveNet model takes approximately $55\times10^{9}$ floating point operations (FLOPs). The activations in any given layer depend on the activations in the previous layer and the previous timestep, so inference must be done one timestep and one layer at a time. A single layer requires only $42\times10^3$ FLOPs, which makes achieving meaningful parallelism difficult. In addition to the compute requirements, the model has approximately $1.6\times 10^6$ parameters, which equate to about 6.4 MB if represented in single precision. (See Appendix~\ref{app:performance-model} for a complete performance model.)

On CPU, a single Haswell or Broadwell core has a peak single-precision throughput of approximately $77\times 10^9$ FLOPs and an L2-to-L1 cache bandwidth of approximately 140 GB/s \footnote{Assuming two 8-wide AVX FMA instructions every cycle and an L2-to-L1 bandwidth of 64 bytes per cycle.}. The model must be loaded from cache once per timestep, which requires a bandwidth of 100 GB/s. Even if the model were to fit in L2 cache, the implementation would need to utilize 70\% of the maximum bandwidth and 70\% of the peak FLOPS in order to do inference in real-time on a single core. Splitting the calculations across multiple cores reduces the difficulty of the problem, but nonetheless it remains challenging as inference must operate at a significant fraction of maximum memory bandwidth and peak FLOPs and while keeping threads synchronized.

A GPU has higher memory bandwidth and peak FLOPs than a CPU but provides a more specialized and hence restrictive computational model. A naive implementation that launches a single kernel for every layer or timestep is untenable, but an implementation based on the persistent RNN technique \cite{diamos2016persistent} may be able to take advantage of the throughput offered by GPUs.

We implement high-speed optimized inference kernels for both CPU and GPU and demonstrate that WaveNet inference at faster-than-real-time speeds is achievable. Table~\ref{tab:inference-speed} lists the CPU and GPU inference speeds for different models. In both cases, the benchmarks include only the autoregressive, high-frequency audio generation and do \emph{not} include the generation of linguistic conditioning features (which can be done in parallel for the entire utterance). Our CPU kernels run at real-time or faster-than-real-time for a subset of models, while the GPU models do not yet match this performance.

\ifdefined\forarxiv\newpage\fi

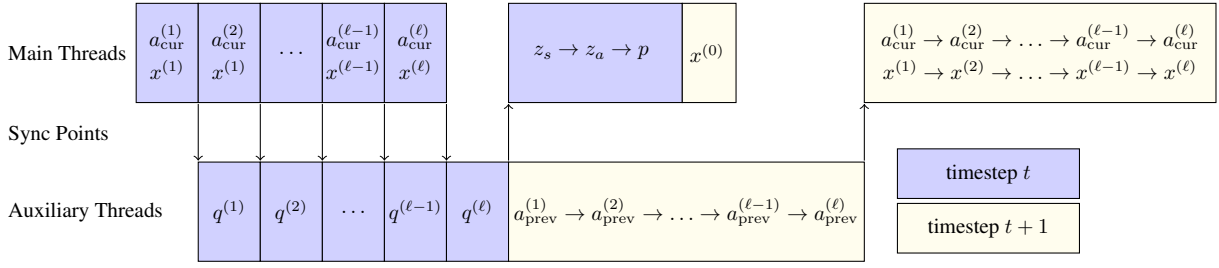
\begin{figure*}[!ht]
    \centering
    \tikzstyle{timestep_t}=[rectangle,fill=blue!18]
    \tikzstyle{timestep_tp1}=[rectangle,fill=yellow!8]
    
    \begin{center}
\begin{tikzpicture}[scale=0.11, every node/.style={scale=0.8}]
\tikzstyle{every node}+=[inner sep=0pt]


\draw (-1,-8.9) node [right] {Main Threads};
\draw (-1,-28.1) node [right] {Auxiliary Threads};
\draw (-1,-18.8) node [right] {Sync Points};

\draw [black,timestep_t] (14.503,-2.9) rectangle (22.003,-14.9);
\draw (18.25,-6.9) node {$a^{(1)}_{\mathrm{cur}}$};
\draw (18.25,-10.9) node {$x^{(1)}$};
\node (A1) at (22.003,-14.9) {};

\draw [black,timestep_t] (22.003,-2.9) rectangle (29.503,-14.9);
\draw (25.75,-6.9) node {$a^{(2)}_{\mathrm{cur}}$};
\draw (25.75,-10.9) node {$x^{(1)}$};
\node (A2) at (29.503,-14.9) {};

\draw [black,timestep_t] (29.503,-2.9) rectangle (37.003,-14.9);
\draw (33.25,-8.9) node {$\hdots$};
\node (A3) at (37.003,-14.9) {};

\draw [black,timestep_t] (37.003,-2.9) rectangle (44.503,-14.9);
\draw (40.75,-6.9) node {$a^{(\ell-1)}_{\mathrm{cur}}$};
\draw (40.75,-10.9) node {$x^{(\ell-1)}$};
\node (A4) at (44.503,-14.9) {};

\draw [black,timestep_t] (44.503,-2.9) rectangle (52.003,-14.9);
\draw (48.25,-6.9) node {$a^{(\ell)}_{\mathrm{cur}}$};
\draw (48.25,-10.9) node {$x^{(\ell)}$};
\node (A5) at (52.003,-14.9) {};

\draw [black,timestep_t] (22.003,-22.1) rectangle (29.503,-34.1);
\draw (25.75,-28.1) node {$q^{(1)}$};
\node (Q1) at (22.003,-22.1) {};

\draw [black,timestep_t] (29.503,-22.1) rectangle (37.003,-34.1);
\draw (33.25,-28.1) node {$q^{(2)}$};
\node (Q2) at (29.503,-22.1) {};

\draw [black,timestep_t] (37.003,-22.1) rectangle (44.503,-34.1);
\draw (40.75,-28.1) node {$\hdots$};
\node (Q3) at (37.003,-22.1) {};

\draw [black,timestep_t] (44.503,-22.1) rectangle (52.003,-34.1);
\draw (48.25,-28.1) node {$q^{(\ell-1)}$};
\node (Q4) at (44.503,-22.1) {};

\draw [black,timestep_t] (52.003,-22.1) rectangle (59.503,-34.1);
\draw (55.75,-28.1) node {$q^{(\ell)}$};
\node (Q5) at (52.003,-22.1) {};

\draw [black,timestep_tp1] (59.503,-22.1) rectangle (102.503,-34.1);
\draw (81,-28.1) node {$a^{(1)}_{\mathrm{prev}} \to a^{(2)}_{\mathrm{prev}} \to \hdots \to a^{(\ell-1)}_{\mathrm{prev}} \to a^{(\ell)}_{\mathrm{prev}}$};
\node (Q6) at (59.503,-22.1) {};
\node (Q7) at (102.503,-22.1) {};


\draw [black,timestep_t] (59.503,-2.9) rectangle (80.503,-14.9);
\draw (69.5,-8.9) node {$z_s \to z_a \to p$};
\node (ZS) at (59.503,-14.9) {};


\draw [black,timestep_tp1] (80.503,-2.9) rectangle (87,-14.9);
\draw (83.5,-8.9) node {$x^{(0)}$};


\draw [black,timestep_tp1] (102.503,-2.9) rectangle (145.003,-14.9);
\draw (123.75,-6.9) node {$a^{(1)}_{\mathrm{cur}} \to a^{(2)}_{\mathrm{cur}} \to \hdots \to a^{(\ell-1)}_{\mathrm{cur}} \to a^{(\ell)}_{\mathrm{cur}}$};
\draw (123.75,-10.9) node {$x^{(1)} \to x^{(2)} \to \hdots \to x^{(\ell-1)} \to x^{(\ell)}$};
\node (ACUR) at (102.503,-14.9) {};

\draw [->] (A1) edge (Q1);
\draw [->] (A2) edge (Q2);
\draw [->] (A3) edge (Q3);
\draw [->] (A4) edge (Q4);
\draw [->] (A5) edge (Q5);
\draw [->] (Q6) edge (ZS);
\draw [->] (Q7) edge (ACUR);

\draw [black,timestep_t] (106.503,-20.5) rectangle (128.503,-26.5);
\draw (117.5,-23.5) node {timestep $t$};
\draw [black,timestep_tp1] (106.503,-27.1) rectangle (128.503,-33.1);
\draw (117.5,-30.1) node {timestep $t+1$};

\end{tikzpicture}
\end{center}
    \caption{Two groups of threads run in parallel. Computation of the $\W{skip}$ is offloaded to the auxiliary threads while the main threads progress through the stack of WaveNet layers. While the main threads are computing the output layer, the auxiliary threads prepare the left $\W{prev}$ half of the WaveNet layer convolutions for the upcoming timestep. Arrows indicate where one thread group waits on results from the other thread group, and are implemented as spinlocks.}
    \label{fig:thread-work-overlay}
\end{figure*}

\subsection{CPU Implementation}

We achieve real-time CPU inference by avoiding any recomputation, doing cache-friendly memory accesses, parallelizing work via multithreading with efficient synchronization, minimizing nonlinearity FLOPs, avoiding cache thrashing and thread contention via thread pinning, and using custom hardware-optimized routines for matrix multiplication and convolution.

For the CPU implementation, we split the computation into the following steps: 
\begin{enumerate}
    \item \textbf{Sample Embedding:} Compute the WaveNet input causal convolution by doing two sample embeddings, one for the current timestep and one for the previous timestep, and summing them with a bias. That is, 
    \begin{align}
        x^{(0)} = \W{emb,prev} \cdot y_{i-1} + \W{emb,cur} \cdot y_{i} + \B{embed}
    \end{align}
    \item \textbf{Layer Inference:} For every layer $j$ from $j=1$ to $\ell$ with dilation width $d$:
    \begin{enumerate}
        \item Compute the left half of the width-two dilated convolution via a matrix-vector multiply:
        \begin{align}
            a^{(j)}_\text{prev} = \W{prev}^{(j)} \cdot x_{i-d}^{(j-1)}
        \end{align}
        \item Compute the right half of the dilated convolution:
        \begin{align}
            a^{(j)}_\text{cur} = \W{cur}^{(j)} \cdot x_i^{(j-1)}
        \end{align}
        \item Compute the hidden state $h^{(j)}$ given the conditioning vector $L^{(j)}_h$:
        \begin{align}
            a^{(j)} &= a^{(j)}_\text{prev} + a^{(j)}_\text{cur} + B^{(j)}_h + L^{(j)}_h \\
            h^{(j)} &= \tanh \left(a^{(j)}_{0:r}\right) \cdot \sigma \left(a^{(j)}_{r:2r}\right),
        \end{align}
        where $v_{0:r}$ denotes the first $r$ elements of the vector $v$ and $v_{r:2r}$ denotes the next $r$ elements. Then, compute the input to the next layer via a matrix-vector multiply:
                \begin{align}
                    x^{(j)} = \W{res}^{(j)} \cdot h^{(j)} + \B{res}^{(j)}
                \end{align}
        \item Compute the contribution to the skip-channel matrix multiply from this layer, accumulating over all layers, with $q^{(0)} = \B{skip}$:
        \begin{align}
            q^{(j)} = q^{(j - 1)} + \W{skip}^{(j)} \cdot h^{(j)}
        \end{align}
    \end{enumerate}
    \item \textbf{Output:} Compute the two output $1\times1$ convolutions:
    \begin{align}
        z_s &= \text{relu}\left(q^{(\ell)}\right) \\
        z_a &= \text{relu}\left(\W{relu} \cdot z_s + \B{relu}\right) \\
        p &= \text{softmax}\left(\W{out} \cdot z_a + \B{out}\right)
    \end{align}
    Finally, sample $y_{i+1}$ randomly from the distribution $p$.
\end{enumerate}

We parallelize these across two groups of threads as depicted in Figure~\ref{fig:thread-work-overlay}. A group of main threads computes $x^{(0)}$, $a^{(j)}_\text{cur}$, $h^{(j)}$, and $x^{(j)}$, $z_a$, and $p$. A group of auxiliary threads computes $a^{(j)}_\text{prev}$, $q^{(j)}$, and $z_s$, with the $a^{(j)}_\text{prev}$ being computed for the next upcoming timestep while the main threads compute $z_a$ and $p$. Each of these groups can consist of a single thread or of multiple threads; if there are multiple threads, each thread computes one block of each matrix-vector multiply, binary operation, or unary operation, and thread barriers are inserted as needed. Splitting the model across multiple threads both splits up the compute and can also be used to ensure that the model weights fit into the processor L2 cache.

Pinning threads to physical cores (or disabling hyperthreading) is important for avoiding thread contention and cache thrashing and increases performance by approximately 30\%.

Depending on model size, the nonlinearities ($\tanh$, $\mathrm{sigmoid}$, and $\mathrm{softmax}$) can also take a significant fraction of inference time, so we replace all nonlinearities with high-accuracy approximations, which are detailed in Appendix \ref{app:nonlinearitydetails}. The maximum absolute error arising from these approximations is $1.5\times 10^{-3}$ for $\tanh$, $2.5\times10^{-3}$ for $\mathrm{sigmoid}$, and $2.4\times10^{-5}$ for $e^x$. With approximate instead of exact nonlinearities, performance increases by roughly $30\%$.

We also implement inference with weight matrices quantized to \texttt{int16} and find no change in perceptual quality when using quantization. For larger models, quantization offers a significant speedup when using fewer threads, but overhead of thread synchronization prevents it from being useful with a larger number of threads.

Finally, we write custom AVX assembly kernels for matrix-vector multiplication using PeachPy \cite{dukhan2015peachpy} specialized to our matrix sizes. Inference using our custom assembly kernels is up to 1.5X faster than Intel MKL and 3.5X faster than OpenBLAS when using \texttt{float32}. Neither library provides the equivalent \texttt{int16} operations.

\subsection{GPU Implementation}

Due to their computational intensity, many neural models are ultimately deployed on GPUs, which can have a much higher computational throughput than CPUs. Since our model is memory bandwidth and FLOP bound, it may seem like a natural choice to run inference on a GPU, but it turns out that comes with a different set of challenges.

Usually, code is run on the GPU in a sequence of kernel invocations, with every matrix multiply or vector operation being its own kernel. However, the latency for a CUDA kernel launch (which may be up to 50 $\mu$s) combined with the time needed to load the entire model from GPU memory are prohibitively large for an approach like this. An inference kernel in this style ends up being approximately 1000X slower than real-time.

To get close to real-time on a GPU, we instead build a kernel using the techniques of persistent RNNs \cite{diamos2016persistent} which generates all samples in the output audio in a single kernel launch. The weights for the model are loaded to registers once and then used without unloading them for the entire duration of inference. Due to the mismatch between the CUDA programming model and such persistent kernels, the resulting kernels are specialized to particular model sizes and are incredibly labor-intensive to write. Although our GPU inference speeds are not quite real-time (Table~\ref{tab:inference-speed}), we believe that with these techniques and a better implementation we can achieve real-time WaveNet inference on GPUs as well as CPUs. Implementation details for the persistent GPU kernels are available in Appendix~\ref{app:gpu-kernels}.

\aftersectionvfill\section{Conclusion}
\label{conclusion}

In this work, we demonstrate that current Deep Learning approaches are viable for all the components of a high-quality text-to-speech engine by building a fully neural system. We optimize inference to faster-than-real-time speeds, showing that these techniques can be applied to generate audio in real-time in a streaming fashion. Our system is trainable without any human involvement, dramatically simplifying the process of creating TTS systems.

Our work opens many new possible directions for exploration. Inference performance can be further improved through careful optimization, model quantization on GPU, and \texttt{int8} quantization on CPU, as well as experimenting with other architectures such as the Xeon Phi. Another natural direction is removing the separation between stages and merging the segmentation, duration prediction, and fundamental frequency prediction models directly into the audio synthesis model, thereby turning the problem into a full sequence-to-sequence model, creating a single end-to-end trainable TTS system, and allowing us to train the entire system with no intermediate supervision. In lieu of fusing the models, improving the duration and frequency models via larger training datasets or generative modeling techniques may have an impact on voice naturalness.
\newpage
\bibliographystyle{icml2016}
\bibliography{references}

\onecolumn
\pagebreak
\appendix
\appendixpage

\section{WaveNet Architecture and Details}
\label{app:wavenet-architecture}

\begin{figure}[ht]
    \centering
    \includegraphics[width=\textwidth]{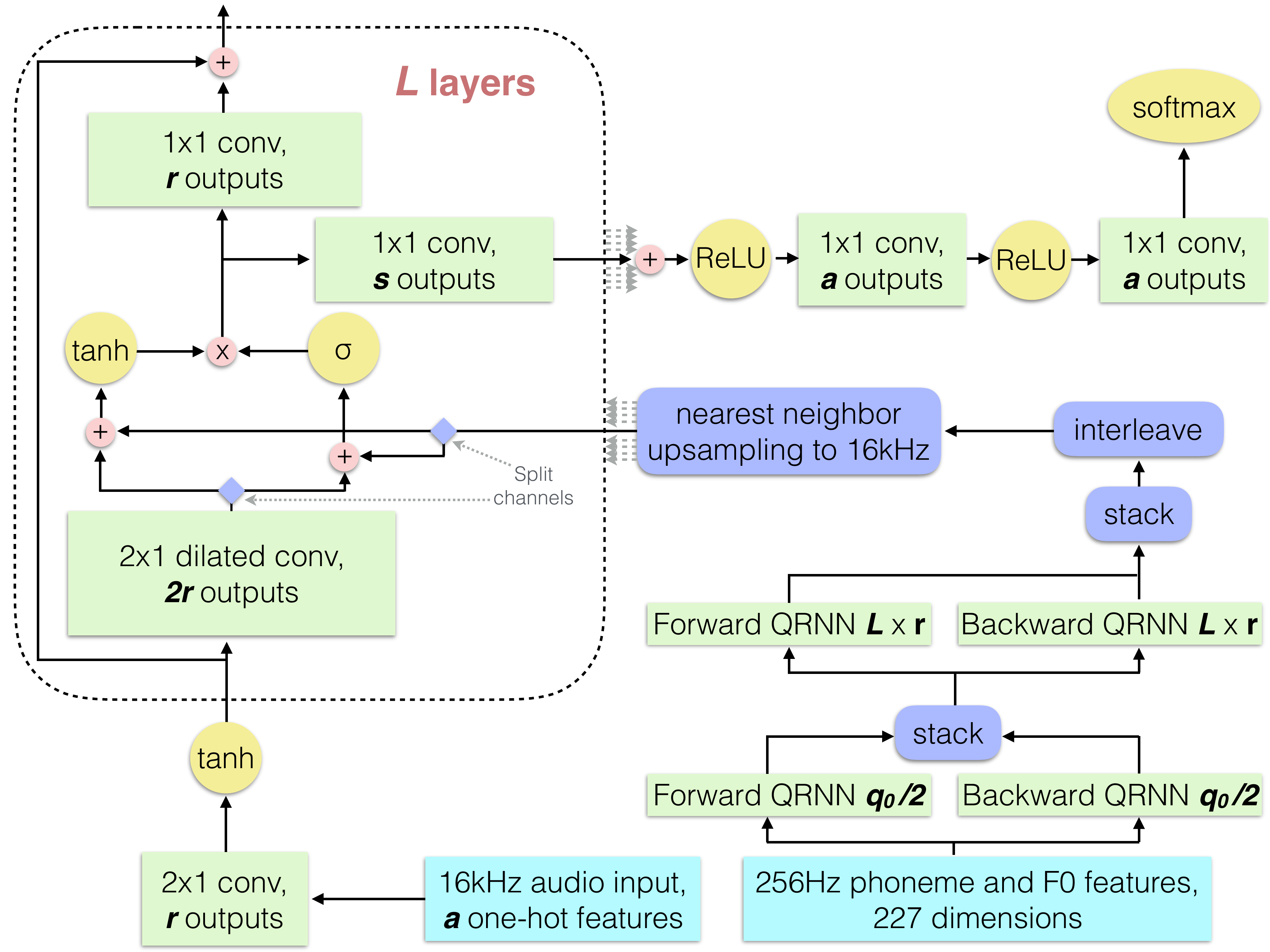}
    \caption{The modified WaveNet architecture. Components are colored according to function: teal inputs, green convolutions and QRNNs, yellow unary operations and softmax, pink binary operations, and indigo reshapes, transposes, and slices.}
    \label{fig:network-sketch}
\end{figure}

The WaveNet consists of a conditioning network $c = C(v)$, which converts low-frequency linguistic features $v$ to the native audio frequency, and an auto-regressive process $P(y_i | c, y_{i-1}, \ldots, y_{i-R})$ which predicts the next audio sample given the conditioning for the current timestep $c$ and a context of $R$ audio samples. $R$ is the receptive field size, and is a property determined by the structure of the network. A sketch of the WaveNet architecture is shown in Figure \ref{fig:network-sketch}. The network details are described in the following subsections.

\subsection{Auto-regressive WaveNet}
\label{app:autoregwavenet}
The structure of the auto-regressive network is parameterized by the number of layers $\ell$, the number of skip channels $s$, and the number of residual channels $r$.

Audio is quantized to $a=256$ values using $\mu$-law companding, as described in Section~2.2 of WaveNet. The one-hot encoded values go through an initial 2x1 convolution which generates the input $x^{(0)} \in \mathbb{R}^r$ for the first layer in the residual stack:
\begin{align}
    x^{(0)} = W_\text{embed} * y + B_\text{embed},
\end{align}
where $*$ is the one-dimensional convolution operator. Since the input audio $y$ is a one-hot vector, this convolution can be done via embeddings instead of matrix multiplies. Each subsequent layer computes a hidden state vector $h^{(i)}$ and then (due to the residual connections between layers) adds to its input $x^{(i-1)}$ to generate its output $x^{(i)}$:
\begin{align}
    h^{(i)} &= \tanh\left(W^{(i)}_h * x^{(i-1)} + B^{(i)}_h + L^{(i)}_h\right) \cdot 
        \sigma\left(W^{(i)}_g * x^{(i-1)} + B^{(i)}_g + L^{(i)}_g\right) \\
    x^{(i)} &= x^{(i-1)} + W^{(i)}_r \cdot h^{(i)} + B^{(i)}_r,
\end{align}
where $L^{(i)}$ is the output for that layer of the conditioning network. Since each layer adds its output to its input, the dimensionality of the layers must remain fixed to the number of residual channels, $r$. Although here this is written as two convolutions, one for $W_h$ and one for $W_g$, it is actually done more efficiently with a single convolution with $r$ input and $2r$ output channels. During inference, this convolution is replaced with two matrix-vector multiplies with matrices $\W{prev}$ (the left half of the convolution) and $\W{cur}$ (the right half). Thus we can reformulate the computation of $h^{(i)}$ for a specific timestep $t$ as follows:
\begin{align}
    h'^{(i)} &= W^{(i)}_{\mathrm{prev}} \cdot x^{(i-1)}_{t-d} + W^{(i)}_{\mathrm{cur}} \cdot x^{(i-1)}_{t} + B^{(i)} + L^{(i)}\\
    h^{(i)} &= \tanh \left(h'^{(i)}_{0:r}\right) \cdot \sigma \left(h'^{(i)}_{r:2r} \right),
\end{align}
where $L^{(i)}$ and is a concatenation of $L^{(i)}_h$ and $L^{(i)}_g$ and $B^{(i)}$ and is a concatenation of $B^{(i)}_h$ and $B^{(i)}_g$.

The hidden state $h^{(i)}$ from each of the layers $1$ through $\ell$ is concatenated and projected with a learned $\W{skip}$ down to the number of skip channels $s$:
\begin{align}
    h &= \begin{bmatrix}h^{(1)} \\ h^{(2)} \\ \vdots \\ h^{(\ell)}\end{bmatrix}\quad & h \in \mathbb{R}^{\ell r} \\
    z_s &= \text{relu}\left(\W{skip} \cdot h + \B{skip}\right), & z_s \in \mathbb{R}^s
\end{align}
where $\text{relu}(x) = \max(0, x)$. 

$z_s$ is then fed through two fully connected relu layers to generate the output distribution $p \in \mathbb{R}^a$:
\begin{align}
    z_a &= \text{relu}\left(W_\text{relu} \cdot z_s + B_\text{relu}\right), & z_a \in \mathbb{R}^a \\
    p &= \text{softmax}\left(W_\text{out} \cdot z_a + B_\text{out}\right)
\end{align}

\subsection{Conditioning Network}

When trained without conditioning information, WaveNet models produce human-like ``babbling sounds", as they lack sufficient long-range information to reproduce words. In order to generate recognizable speech, every timestep is conditioned by an associated set of linguistic features. This is done by biasing every layer with a per-timestep conditioning vector generated from a lower-frequency input signal containing phoneme, stress, and fundamental frequency features.

The frequency of the audio is significantly higher than the frequency of the linguistic conditioning
information, so an upsampling procedure is used to convert from lower-frequency linguistic features
to higher-frequency conditioning vectors for each WaveNet layer.

The original WaveNet does upsampling by repetition or through a transposed convolution. Instead, we first pass our input features through two bidirectional quasi-RNN layers \cite{bradbury2016quasi} with $\mathrm{fo}$-pooling and 2x1 convolutions. A unidirectional QRNN layer with $\mathrm{fo}$-pooling is defined by the following equations:
\begin{align}
    \tilde h &= \tanh\left(W_h * x + B_h\right) \\
    o &= \sigma\left(W_o * x + B_o\right) \\
    f &= \sigma\left(W_f * x + B_f\right) \\
    h_t &=  f_t \cdot h_{t-1} + (1 - f_t) \cdot \tilde h_t  \\
    z_t &= o_t \cdot h_t
\end{align}
A bidirectional QRNN layer is computed by running two unidirectional QRNNs, one on the input sequence and one on a reversed copy of the input sequence, and then stacking their output channels. After both QRNN layers, we interleave the channels, so that the $\tanh$ and the $\mathrm{sigmoid}$ in the WaveNet both get channels generated by the forward QRNN and backward QRNN.

Following the bidirectional QRNN layers, we upsample to the native audio frequency by repetition\footnote{Upsampling using bilinear interpolation slowed convergence and reduced generation quality by adding noise or causing mispronunciations, while bicubic upsampling led to muffled sounds. Upsampling by repetition is done by computing the ratio of the output frequency to the input frequency and repeating every element in the input signal an appropriate number of times.}.

We find that the model is very sensitive to the upsampling procedure: although many variations of the conditioning network converge, they regularly produce phoneme mispronunciations.

\subsection{Input Featurization}

Our WaveNet is trained with 8-bit $\mu$-law companded audio which is downsampled to 16384 Hz from 16-bit dual-channel PCM audio at 48000 Hz. It is conditioned on a 256 Hz phoneme signal. The conditioning feature vector has 227 dimensions. Of these two are for fundamental frequency. One of these indicates whether the current phoneme is voiced (and thus has an F0) and the other is normalized log-frequency, computed by normalizing the log of F0 to minimum observed F0 to be approximately between -1 and 1. The rest of the features describe the current phoneme, the two previous phonemes, and the two next phonemes, with each phoneme being encoded via a 40-dimensional one-hot vector for phoneme identity (with 39 phonemes for ARPABET phonemes and 1 for silence) and a 5-dimensional one-hot vector for phoneme stress (no stress, primary stress, secondary stress, tertiary stress, and quaternary stress). Not all of the datasets we work with have tertiary or quaternary stress, and those features are always zero for the datasets that do not have those stress levels.

In our experiments, we found that including the phoneme context (two previous and two next phonemes) is crucial for upsampling via transposed convolution and less critical but still important for our QRNN-based upsampling. Although sound quality without the phoneme context remains high, mispronunciation of a subset of the utterances becomes an issue. We also found that including extra prosody features such as word and syllable breaks, pauses, phoneme and syllable counts, frame position relative to phoneme, etc, were unhelpful and did not result in higher quality synthesized samples.

In order to convert from phonemes annotated with durations to a fixed-frequency phoneme signal, we sample the phonemes at regular intervals, effectively repeating each phoneme (with context and F0) a number proportional to its duration. As a result, phoneme duration is effectively quantized to $\nicefrac{1}{256} \text{ sec} \approx 4\text{ms}$.

We use Praat \cite{boersma2002praat} in batch mode to compute F0 at the appropriate frequency, with a minimum F0 of 75 and a maximum F0 of 500. The Praat batch script used to generate F0 is available at
\ifdefined\isaccepted
    \href{https://github.com/baidu-research/deep-voice/blob/master/scripts/f0-script.praat}{https://github.com/baidu-research/deep-voice/blob/master/scripts/f0-script.praat}
\else
\href{https://github.com/XXX-anonymous-for-review-XXX/praat-f0-script}{https://github.com/XXX-anonymous-for-review-XXX/praat-f0-script}
\fi and can be run with \texttt{praat --run f0-script.praat}.

\subsection{Sampling from Output Distribution}

At every timestep, the synthesis model produces a distribution over samples, $P(s)$, conditioned on the previous samples and the linguistic features. In order to produce the samples, there are a variety of ways you could choose to use this distribution:

\begin{itemize}
    \item \textbf{Direct Sampling:} Sample randomly from $P(y)$.
    \item \textbf{Temperature Sampling:} Sample randomly from a distribution adjusted by a temperature $t$
    \begin{align}
    \tilde P_t(y) = \frac{1}{Z}P(y)^{\nicefrac{1}{t}},
    \end{align}
    where $Z$ is a normalizing constant.
    \item \textbf{Mean:} Take the mean of the distribution $\mathrm{E}_P[y]$.
    \item \textbf{Mode:} Take the most likely sample, $\text{argmax}\;P(y).$
    \item \textbf{Top $k$:} Sample from an adjusted distribution that only permits the top $k$ samples
        \begin{align}
        \tilde P_k(y) = \begin{cases}
            0 & \text{if } y < k\text{th}(P(y)) \\
            P(y) / Z & \text{otherwise},
        \end{cases}
        \end{align} where $Z$ is a normalizing constant.
\end{itemize}

We find that out of these different sampling methods, only direct sampling produces high quality
outputs. Temperature sampling produces acceptable quality results, and indeed outperforms direct
sampling early on in training, but for converged models is significantly worse. This observation
indicates that the generative audio model accurately learns a conditional sample distribution and
that modifying this distribution through the above heuristics is worse than just using the learned
distribution.

\subsection{Training}

We observed several tendencies of the models during training. As expected, the randomly initialized
model produces white noise. Throughout training, the model gradually increases the signal to noise
ratio, and the volume of the white noise dies down while the volume of the speech signal increases.
The speech signal can be inaudible for tens of thousands of iterations before it dominates the white
noise.

In addition, because the model is autoregressive, rare mistakes can produce very audible
disturbances. For example, a common failure mode is to produce a small number of incorrect samples
during sampling, which then results in a large number incorrect samples due to compounding errors.
This is audible as a brief period of loud noise before the model stabilizes. The likelihood of this
happening is higher early on in training, and does not happen in converged models.

\section{Phoneme Model Loss}
\label{app:phonememodel}
The loss for the $n^{th}$ phoneme is
\begin{align}
    L_n = |\hat{t}_n - t_n| + \lambda_1\text{CE}(\hat{p}_n, p_n) + \lambda_2\sum_{t=0}^{T-1}|\widehat{F0}_{n,t} - F0_{n,t}| + \lambda_3\sum_{t=0}^{T-2}|\widehat{F0}_{n,t+1} - \widehat{F0}_{n,t}|,
\end{align}
where $\lambda_i$'s are tradeoff constants, $\widehat{t}_n$ and $t_n$ are the estimated and ground-truth durations of the $n^{th}$ phoneme, $\hat{p}_n$ and $p_n$ are the estimated and ground-truth probabilities that the $n^{th}$ phoneme is voiced, $\text{CE}$ is the cross-entropy function, $\widehat{F}_{n,t}$ and $F_{n,t}$ are the estimated and ground-truth values of the fundamental frequency of the $n^{th}$ phoneme at time $t$. $T$ time samples are equally spaced along the phoneme duration.

\section{Nonlinearity Approximation Details}

During inference, we replace exact implementations of the neural network nonlinearities with high-accuracy rational approximations. In this appendix, we detail the derivation of these approximations.

\label{app:nonlinearitydetails}
\subsection{tanh and sigmoid approximation}
Denoting $\tilde{e}(x)$ as an approximation to $e^{|x|}$, we use the following approximations for $\tanh$ and $\sigma$:
\begin{align}
\label{eqn:tanhapp}
\tanh(x) & \approx \text{sign}(x) \frac{ \tilde{e}(x) - \nicefrac{1}{\tilde{e}(x)}}{\tilde{e}(x) + \nicefrac{1}{\tilde{e}(x)}} \\
\label{eqn:sigapp}
\sigma(x) &  \approx \begin{cases} 
        \frac{\tilde{e}(x)}{1 + \tilde{e}(x)} & x \geq 0 \\
        \frac{1}{1 + \tilde{e}(x)} & x \leq 0
    \end{cases}
\end{align}
We choose a forth-order polynomial to represent $\tilde{e}(x)$. The following fit produces accurate values for both $\tanh(x)$ and $\sigma(x)$:
\begin{align}
\tilde{e}(x) = 1 + |x| + 0.5658 x^2 + 0.143 x^4
\end{align}
By itself, $\tilde{e}(x)$ is not a very good approximate function for $e^{|x|}$, but it yields good approximations when used to approximate $\tanh$ and $\sigma$ as described in Equations \ref{eqn:tanhapp} and \ref{eqn:sigapp}.

\subsection{$e^x$ approximation}
We follow the approach of \cite{IanStephensonBook} to calculate an approximate $e^x$ function. Instead of approximating $e^x$ directly, we approximate $2^x$ and use the identity $e^x = 2^{\nicefrac{x}{\ln 2}}$. 

Let $\floor{x}$ to be the floor of $x\in \mathbb{R}$. Then,
\begin{align*}
    2 ^ x &=  2^{\floor{x}} \cdot 2^{x - \floor{x}} \\
    &= 2^{\floor{x}} \cdot \left( 1 + (2 ^ {x - \floor{x}} - 1) \right)
\end{align*}
where $0 \leq 2 ^ {x - \floor{x}} - 1 < 1$ since $0 \le x - \floor{x} < 1$. If we use a $32$-bit float to represent  $2 ^ x$, then $\floor{x} + 127$ and $2 ^ {x - \floor{x}} - 1$ are represented by  the exponent and fraction bits of $2^x$. Therefore, if we interpret the bytes pattern of $2 ^ x$ as a $32$-bits integer (represented by $I_{2^x}$), we have
\begin{equation}
    \label{eqn:I2x}
    I_{2^x} = (\floor{x} + 127) \cdot 2 ^ {23} + (2 ^ {x - \floor{x}} - 1) \cdot 2 ^ {23}.
\end{equation}
Rearranging the Equation \ref{eqn:I2x} and using $z = x - \floor{x}$ results to
\begin{equation}
    \label{eqn:I2xrearranged}
    I_{2^x} = \left( x + 126 + \{2^ z - z\} \right) \cdot 2 ^ {23}    
\end{equation}
If we can accurately approximate $g(z)=2^z - z$ over $z\in[0, 1)$, then interpreting back the byte representation of $I_{2^x}$ in Equation \ref{eqn:I2xrearranged} as a $32$-bits float, we can 
accurately approximate $2 ^ x$. We use a rational approximation as
\begin{align}
    g(z) \approx -4.7259162 + \frac{27.7280233}{4.84252568 - z} - 1.49012907z,
\end{align}
which gives are maximum error $2.4\times 10^{-5}$ for $x\in(-\infty, 0]$.

\section{Persistent GPU Kernels}
\label{app:gpu-kernels}
A NVIDIA GPU has multiple Streaming Multiprocessors (SMs), each of which has a register file and a L1 cache. There is also a coherent L2 cache that is shared by all SMs. The inference process needs to generate one sample every 61 $\mu$s. Due to the high latency of a CUDA kernel launch and of reading small matrices from GPU memory, the entire audio generation process must be done by a single kernel with the weights loaded into the register file across all SMs. This raises two challenges---how to split the model across registers in a way to minimize communication between SMs and how to communicate between SMs given the restrictions imposed by the CUDA programming model.

We split the model across the register file of 24 SMs, numbered SM1 $\cdot$ SM24, of a TitanX GPU. We do not use SM24. SM1 to SM20 store two adjacent layers of the residual stack. This means SM1 stores layers 1 and 2, SM2 stores layers 3 and 4 and so on and so forth. Each layer has three matrices and three bias vectors---$W_{\mathrm{prev}}$, $B_{\mathrm{prev}}$, $W_{\mathrm{cur}}$, $B_{\mathrm{cur}}$, that are for the dilated convolutions and $W_r$, $B_r$. Thus SM$i$ generates two hidden states $h^{(2i)}$ and $h^{(2i+1)}$ and an output $x^{(2i)}$. Each SM also stores the rows of the $W_{\mathrm{skip}}$ matrix that will interact with the generated hidden state vectors. Thus $W_{\mathrm{skip}}$ is partitioned across 20 SMs. Only SM20 needs to store $B_{\mathrm{skip}}$. SM21 stores $W_{\mathrm{relu}}$ and $B_{\mathrm{relu}}$. Finally, $W_{\mathrm{out}}$ is split across two SMs---SM22 and SM23 because of register file limitations and SM23 stores $B_{\mathrm{out}}$.

The next challenge is to coordinate the data transfer between SMs, since the CUDA programming model executes one kernel across all SMs in parallel. However we want execution to go sequentially in a round robin fashion from SM1 to SM23 and back again from SM1 as we generate one audio sample at a time. We launch our CUDA kernel with 23 thread blocks and simulate such sequential execution by spinning on locks, one for each SM, that are stored in global memory and cached in L2. First SM1 executes two layers of the WaveNet model to generate $h^{(1)}$, $h^{(2)}$ and $x^{(2)}$. It then  unlocks the lock that SM2 is spinning on and sets its own lock. It does this by bypassing the L1 cache to write to global memory so that all SMs have a coherent view of the locks. Then SM2 does the same for SM3 and this sequential locking and unlocking chain continues for each SM. Finally SM23 generates the output distribution $p$ for timestep $t$ and unlocks SM1 so that entire process can repeat to generate $p$ for timestep $t+1$.

Just like locks, we pass data between SMs, by reading and writing to global memory by bypassing the L1 cache. Since NVIDIA GPUs have a coherent L2 cache, a global memory write bypassing the L1, followed by a memory fence results in a coherent view of memory across SMs. 

This partitioning scheme however is quite inflexible and only works for specific values of $l$, $r$ and $s$ shown in Table~\ref{tab:inference-speed}. This is because each SM has a fixed sized register file and combined with the relatively inflexible and expensive communication mechanism between SMs implies that splitting weight matrices between SMs is challenging. Any change in those parameters means a new kernel has to be written, which is a very time consuming process.

There are two main reasons why the GPU kernels are slower than CPU kernels. Firstly, synchronization between SMs in a GPU is expensive since it is done by busy waiting on locks in L2 cache. Secondly even though we divide the model in a way that will fit in the register file of each SM, the CUDA compiler still spills to L1 cache. We hope that with handcrafted assembly code, we will be able to match the performance of CPU kernels. However, the lack of parallelism in WaveNet inference makes it difficult to hide the latencies inherent in reading and writing small matrices from GPU memory which are exposed in the absence of a rich cache hierarchy in GPUs.

\section{Performance model}
\label{app:performance-model}

We present a performance model for the autoregressive WaveNet architecture described in Appendix~\ref{app:autoregwavenet}. In our model a dot product between two vectors of dimension $r$ takes $2r$ FLOPS---$r$ multiplications and $r$ additions. This means that a matrix-vector multiply between $W$, an $r \times r$ matrix and $x$, a $r \times 1$ vector takes $2r \cdot r = 2r^2$ FLOPs. Thus calculating $h'^{(i)}$ uses
\begin{align}
    \text{Cost}\left(h'^{(i)}\right) = (2r \cdot 2r) + (2r \cdot 2r) + 2r + 2r + 2r \;\; \mathrm{FLOPs}
\end{align}

Let division and exponentiaton take $f_d$ and $f_e$ FLOPs respectively. This means $\tanh$ and $\sigma$ takes $(f_d + 2f_e + 1)$ FLOPs. Thus calculating $h^{(i)}$ takes $2r \cdot \left(f_d + 2f_e + 1\right) + r$ FLOPs. Finally calculating $x^{(i)}$ for each layer takes $r + (2r \cdot r) + r$ FLOPs. This brings the total FLOPs for calculating one layer to
\begin{align}
    \text{Cost}\left(\text{layer}\right) = 10r^2 + 11r + 2r(f_d + f_e) \;\; \mathrm{FLOPs}
\end{align}

Under the same model, calculating $z_s$ takes $(\ell \cdot 2r) \cdot s + s + s$ FLOPs, where we assume that $\mathrm{relu}$ takes 1 FLOP. Similarly, calculating $z_a$ takes $2s \cdot a + a + a$ FLOPs and $W_{\mathrm{out}} \cdot z_a + B_{\mathrm{out}}$ takes $2a \cdot a + a$ FLOPs.

Calculating the numerically stable $\mathrm{softmax}$ takes one $\mathrm{max}$, one subtract, one exponentiation, one sum and one division per element of a vector. Hence calculating $p$ takes $3a + a(f_d + f_e)$ FLOPs.

Adding it all up, our final performance model to generate each audio sample is as follows:
\begin{align}
    \text{Cost}\left(\text{sample}\right) =\ell \left( 10r^2 + 11r + 2r(f_d + f_e) \right) + s(2r \cdot \ell + 2) + a(2s + 2a + 3) + a(3 + f_d + f_e) \;\; \mathrm{FLOPS}
\end{align}

If we let $\ell=40$, $r=64$, and $s=a=256$, and assume that $f_d = 10$ and $f_e = 10$, with a sampling frequency of 16384Hz, we have approximately $55\times10^9$ FLOPs for every second of synthesis. 

\end{document}